\documentclass[10pt,twocolumn,letterpaper]{article}

\usepackage{iccv}
\usepackage{times}
\usepackage{epsfig}
\usepackage{graphicx}
\usepackage{amsmath}
\usepackage{amssymb}

\usepackage{verbatim}
\usepackage{array}

\newcommand{\name}{SACANet}
\usepackage{graphicx}

\usepackage[breaklinks=true,bookmarks=false]{hyperref}

\iccvfinalcopy 

\def\iccvPaperID{5} 

\ificcvfinal\fi

\begin{document}

\title{
	Crowd Counting on Images with Scale Variation and
	Isolated Clusters
}

\author{
	Haoyue Bai\quad 
	Song Wen\quad 
	S.-H. Gary Chan\\
	Department of Computer Science and Engineering\\
	The Hong Kong University of Science and Technology, Hong Kong, China\\
	{\tt\small \{hbaiaa, swenaa, gchan\}@cse.ust.hk}
}

\maketitle
\ificcvfinal\thispagestyle{empty}\fi


\begin{abstract}
	Crowd counting is to estimate the number of objects (e.g., people or vehicles) in an image of unconstrained congested scenes.
	Designing a general crowd counting algorithm applicable to a wide range of crowd images is challenging, mainly due to the possibly large variation in object scales and the presence of many isolated small clusters.  
	Previous approaches based on convolution operations with multi-branch architecture are effective for only some narrow bands of scales, and have not captured the long-range contextual relationship due to isolated clustering.
	To address that, we propose \name{}, a novel {\bf s}cale-{\bf a}daptive long-range {\bf c}ontext-{\bf a}ware {\bf net}work  for crowd counting.
	
	\name{} consists of three major modules: the pyramid contextual module which extracts long-range contextual information and enlarges the receptive field, a scale-adaptive self-attention multi-branch module to attain high scale sensitivity and detection accuracy of isolated clusters, and a hierarchical fusion module to fuse multi-level self-attention features.
	With group normalization, \name{} achieves better optimality in the training process.
	We have conducted extensive experiments using the VisDrone2019 People dataset, the VisDrone2019 Vehicle dataset, and some other challenging benchmarks. As compared with the state-of-the-art methods, \name{} is shown to be effective, especially for extremely crowded conditions with diverse scales and scattered clusters, and achieves much lower MAE as compared with baselines.
\end{abstract}



\section{Introduction}

Crowd counting is to estimate the number of objects (such as people or
vehicles) in unconstrained congested environments, where the image 
is often taken by a surveillance camera or unmanned aerial
vehicle (UAV).
Crowd counting has attracted widespread attention due to its application in public safety, congestion monitoring and traffic management~\cite{sindagi2018survey},~\cite{kang2018beyond}.

A promising approach for
crowd counting is to use density map regression-based Convolutional Neural Networks (CNNs), which estimate the number of objects per unit pixel instead of detecting, recognizing and counting objects in the whole image.
Despite recent advances, precise crowd counting remains challenging. This is mainly due to the following two factors:

\newcolumntype{C}[1]{>{\centering\let\newline\\\arraybackslash\hspace{0pt}}m{#1}}

\begin{table*}
	\centering
	\vspace{-0.15in}
	\caption{Illustration of scale variation and small isolated clusters based on three typical images from the VisDrone2019 Vehicle dataset.}
	\begin{tabular}{|C{0.2\columnwidth}|c|c|c|}
		\hline
		~ &Image $A$ &Image $B$ &Image $C$ \\		
		\hline\hline
		&&&\\                
		Original image &\includegraphics[width=0.56\columnwidth, height=0.135\textwidth]{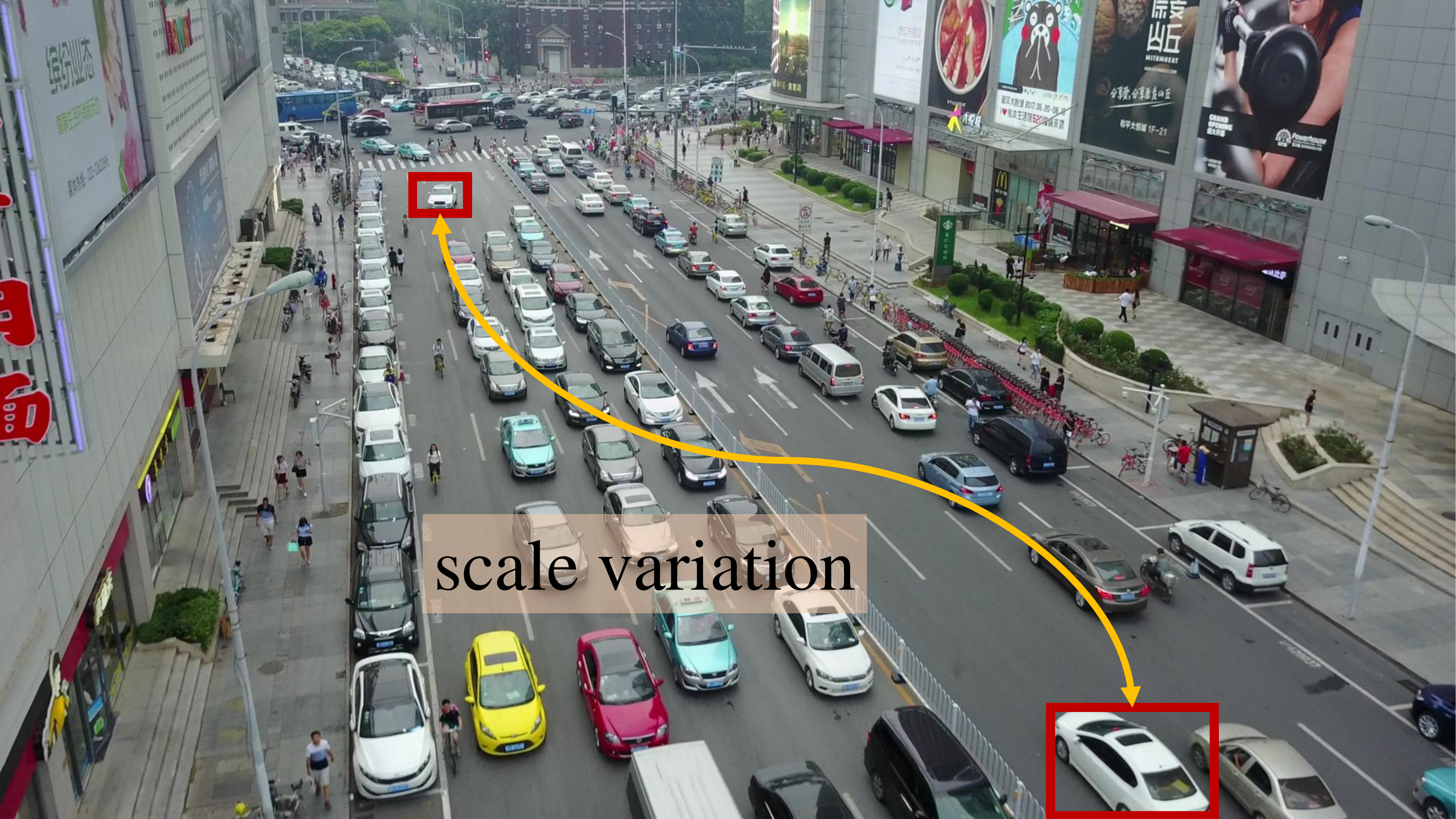} &\includegraphics[width=0.56\columnwidth, height=0.135\textwidth]{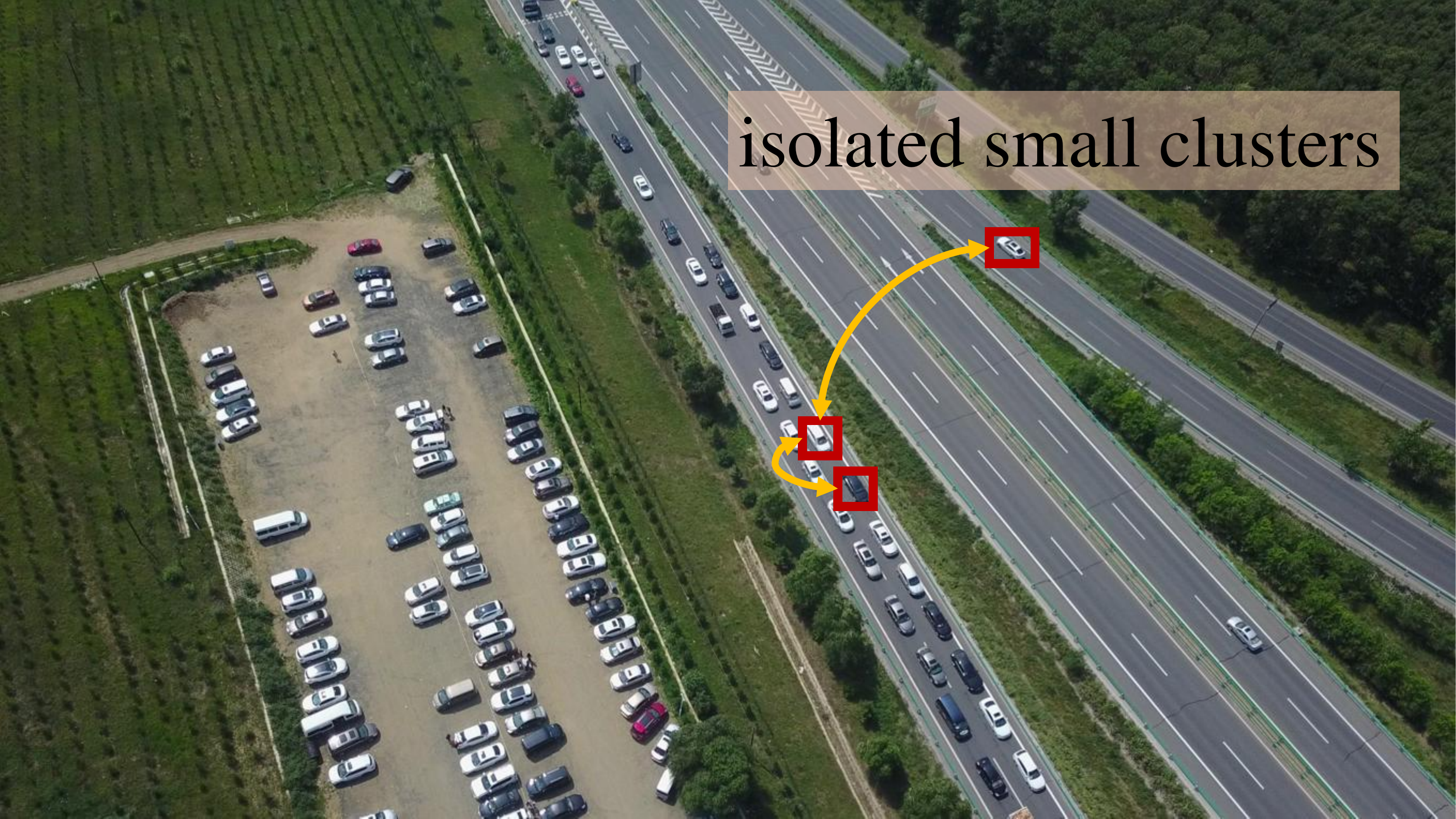} &\includegraphics[width=0.56\columnwidth, height=0.135\textwidth]{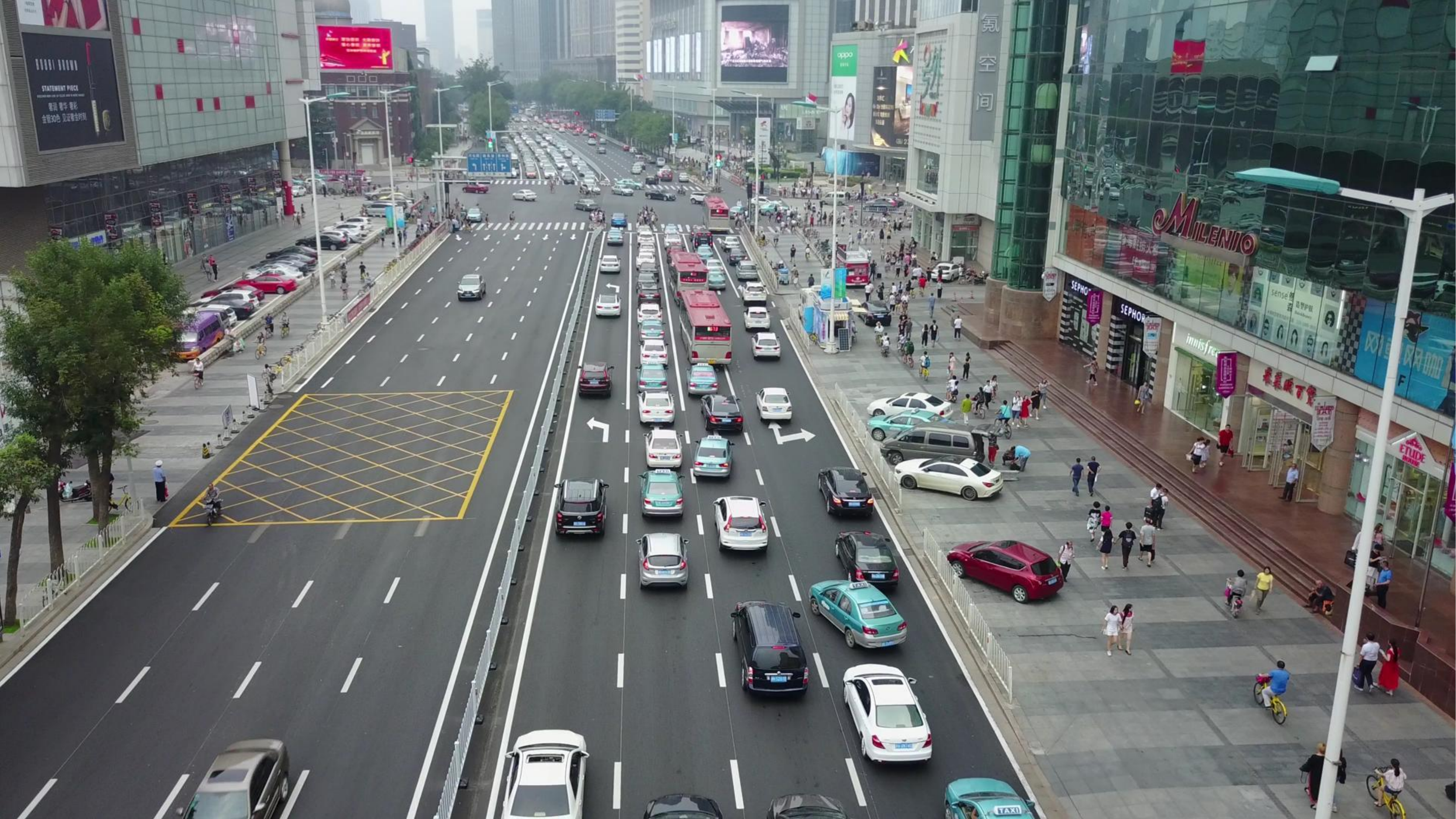} \\
		\hline
		&&&\\                
		Density  map &\includegraphics[width=0.56\columnwidth, height=0.135\textwidth]{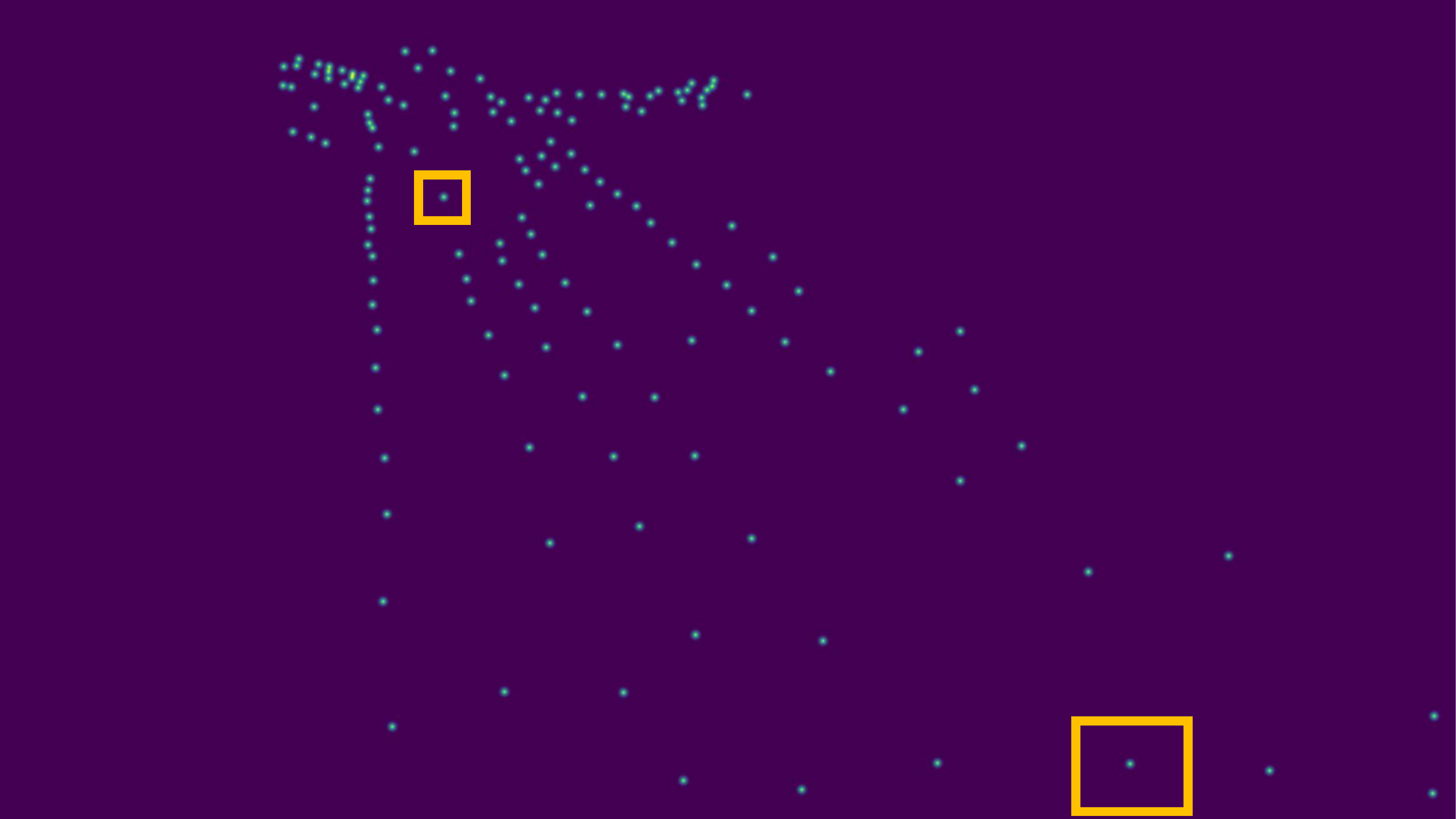} &\includegraphics[width=0.56\columnwidth, height=0.135\textwidth]{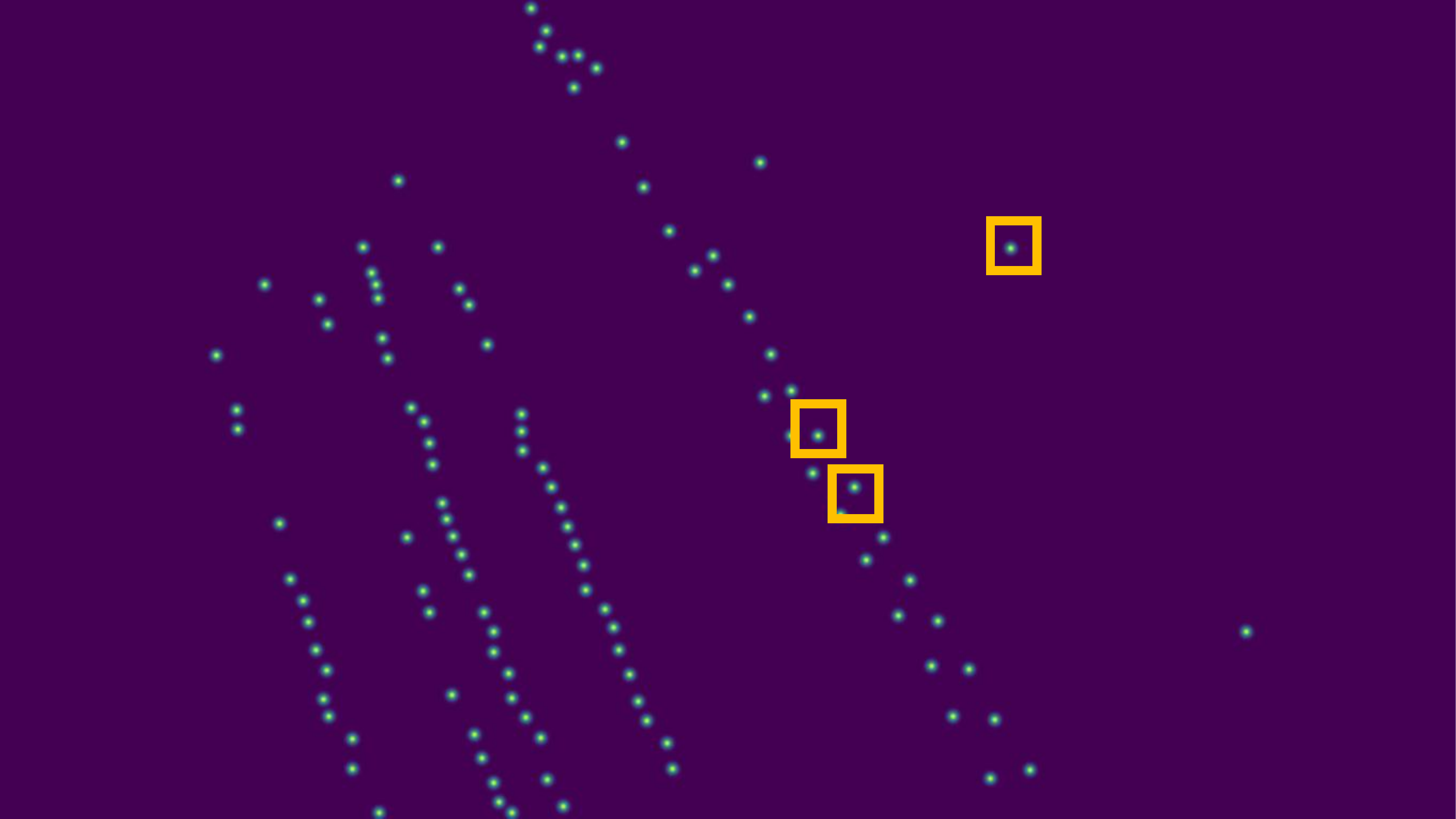} &\includegraphics[width=0.56\columnwidth, height=0.135\textwidth]{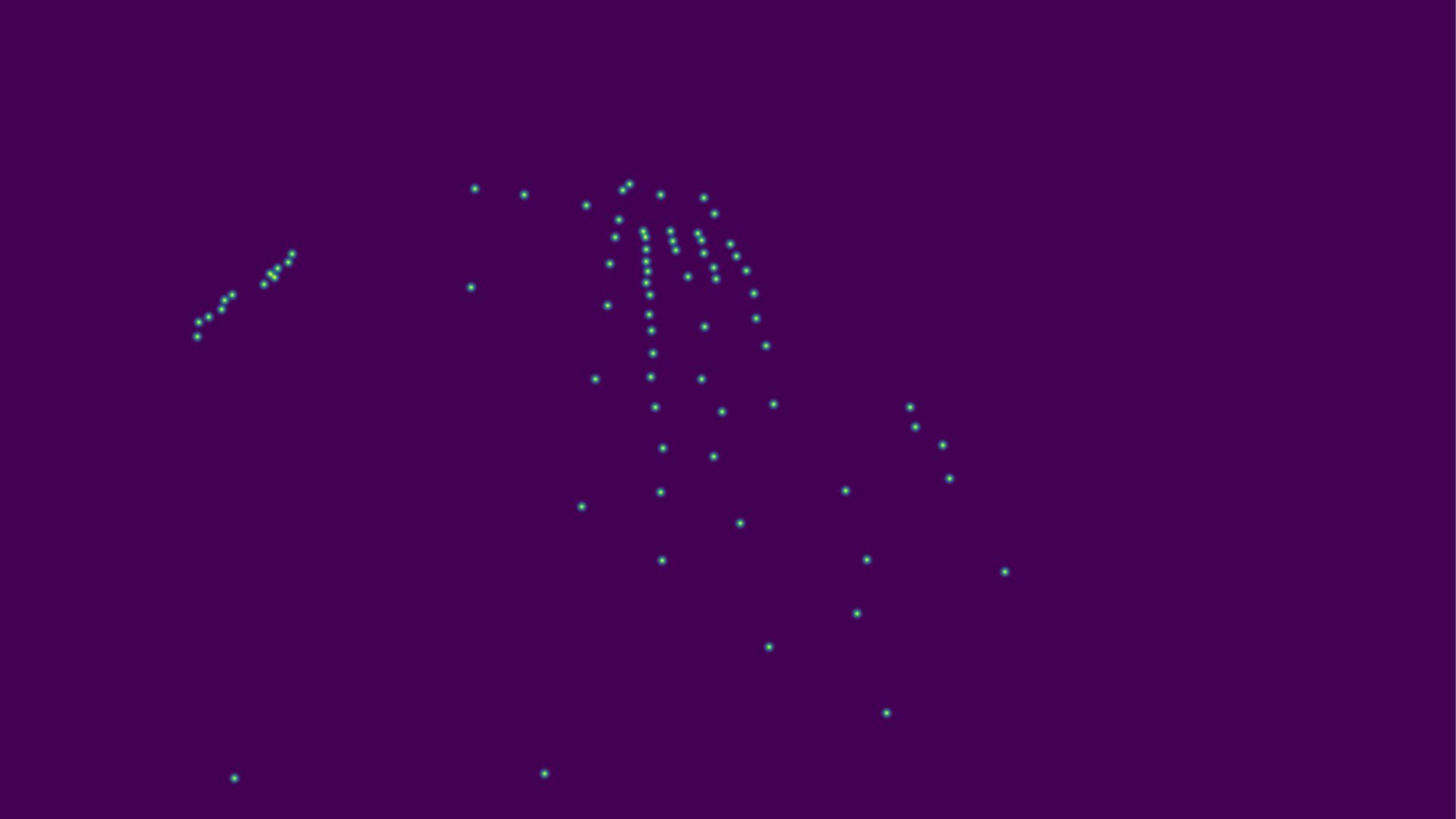} \\
		\hline
		&&&\\                                
		Scale  distribution &\includegraphics[width=0.56\columnwidth, height=0.19\textwidth]{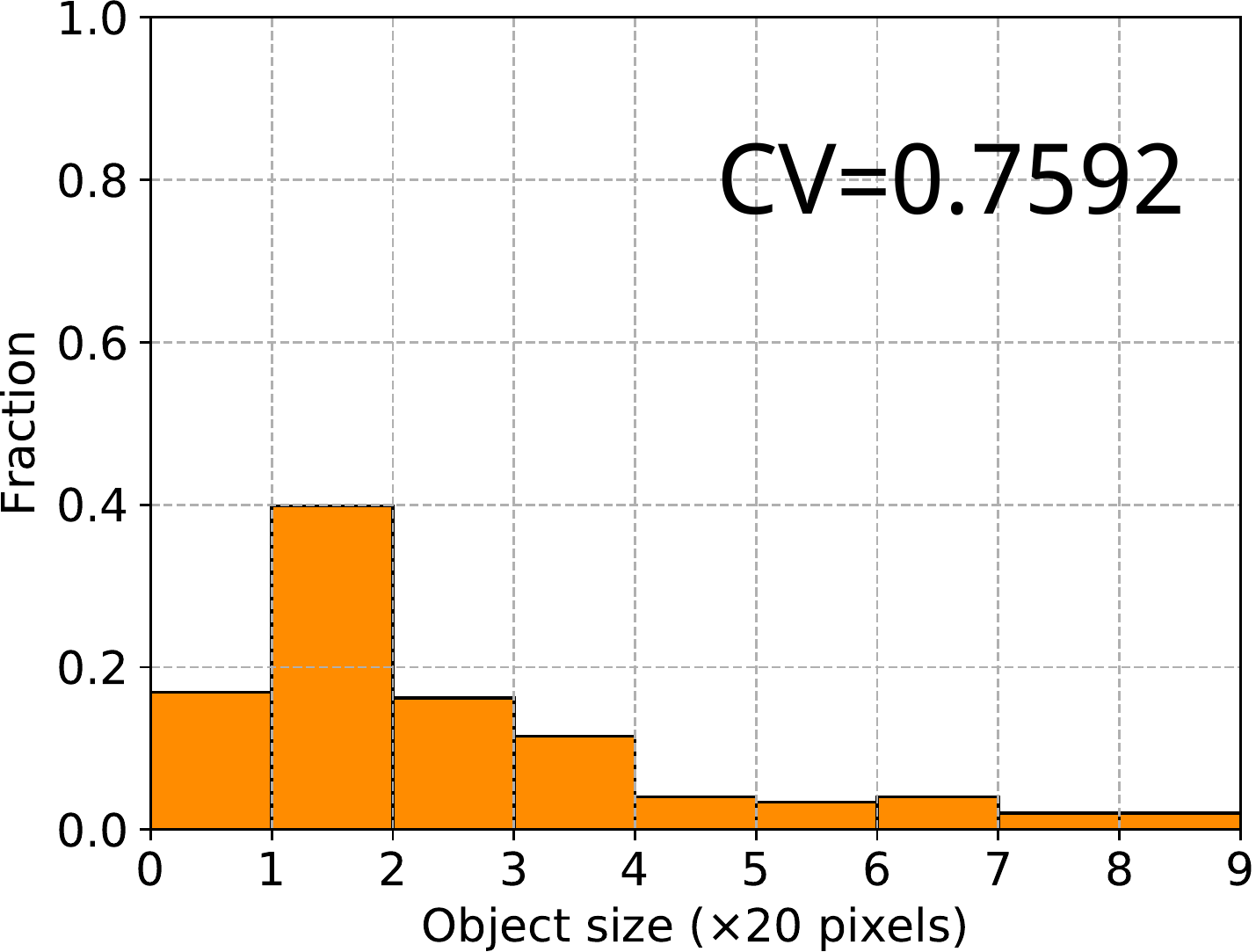} &\includegraphics[width=0.56\columnwidth, height=0.19\textwidth]{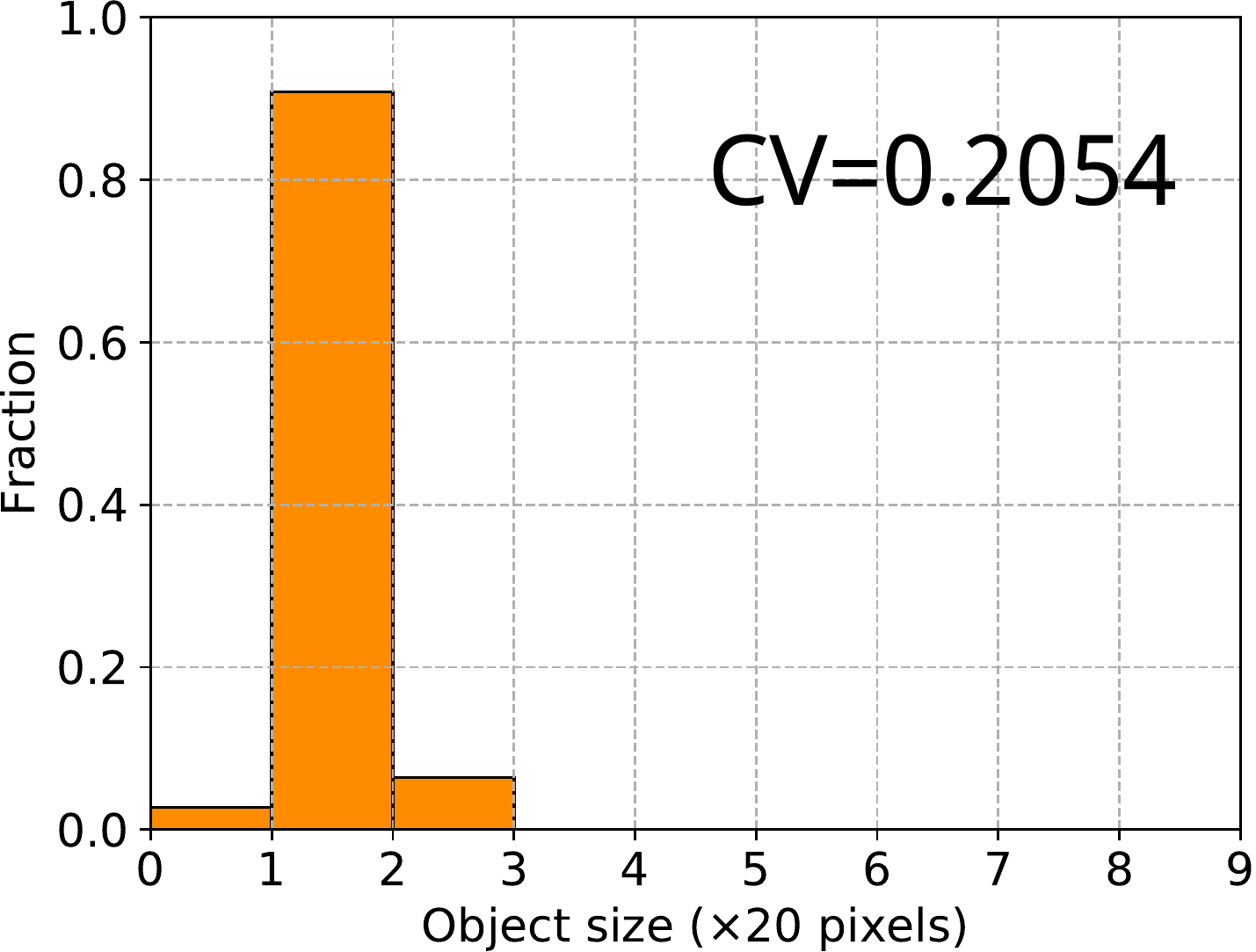} &\includegraphics[width=0.56\columnwidth, height=0.19\textwidth]{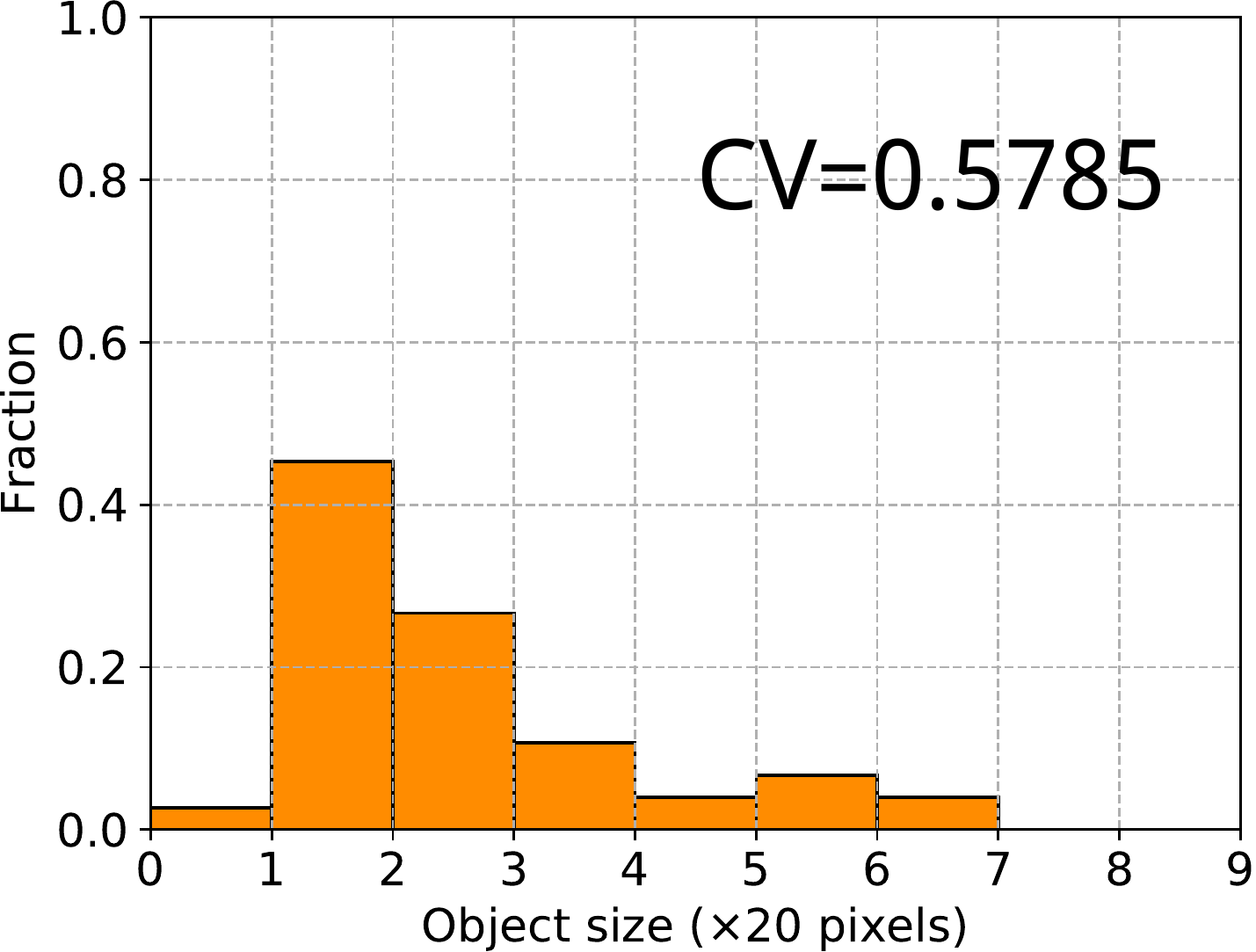} \\
		
		\hline
		&&&\\     
		Distance  distribution &\includegraphics[width=0.56\columnwidth, height=0.19\textwidth]{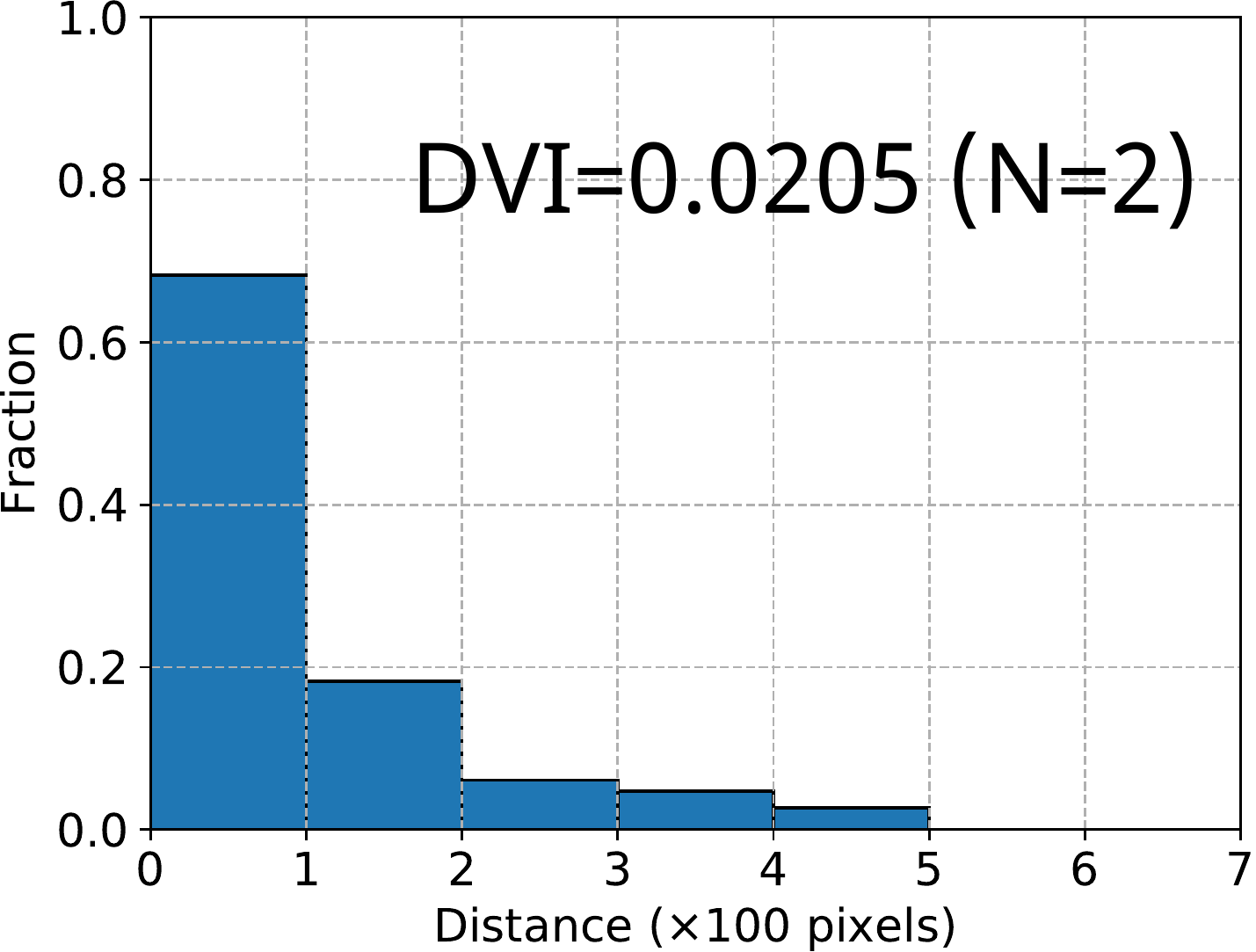} &\includegraphics[width=0.56\columnwidth, height=0.19\textwidth]{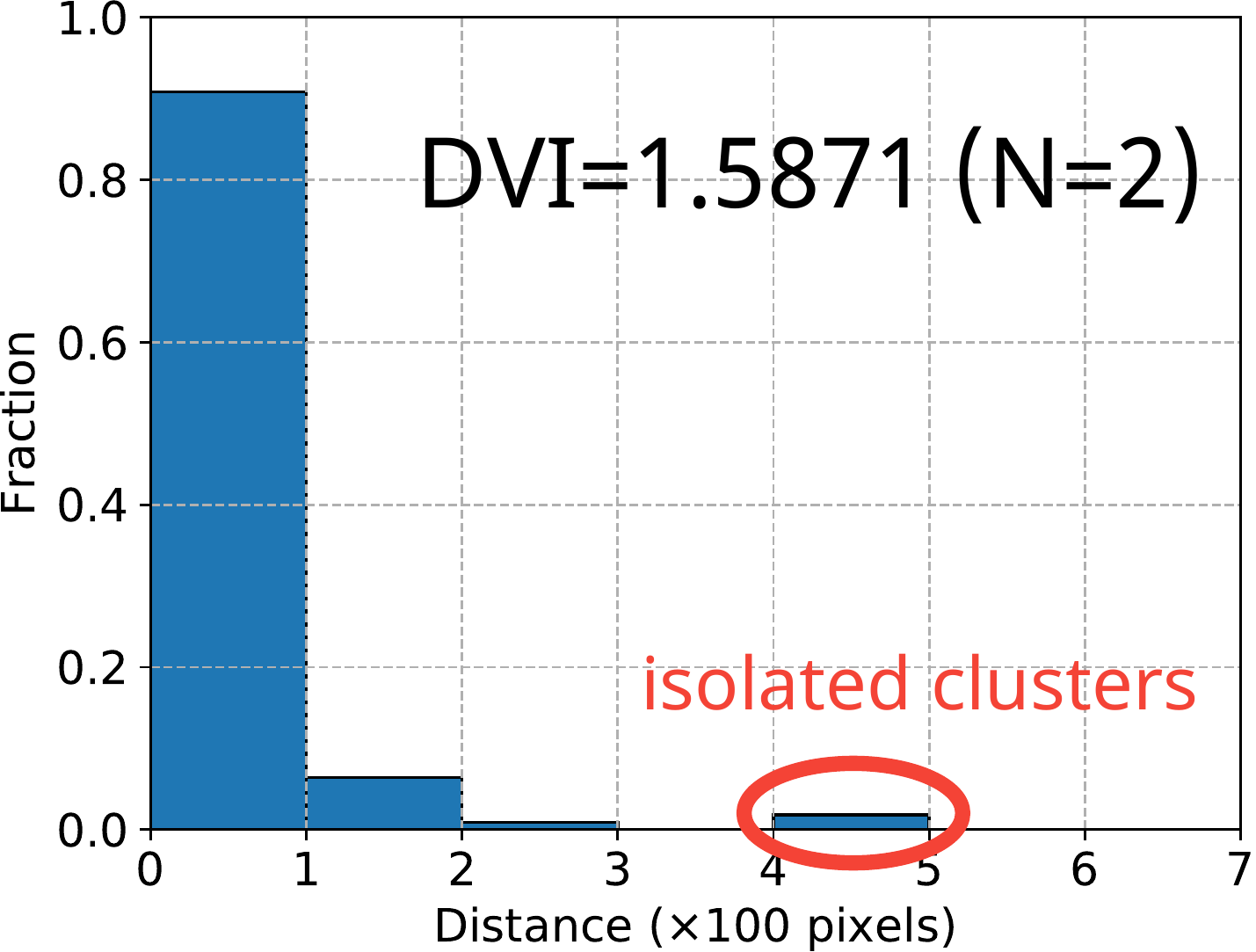} &\includegraphics[width=0.56\columnwidth, height=0.19\textwidth]{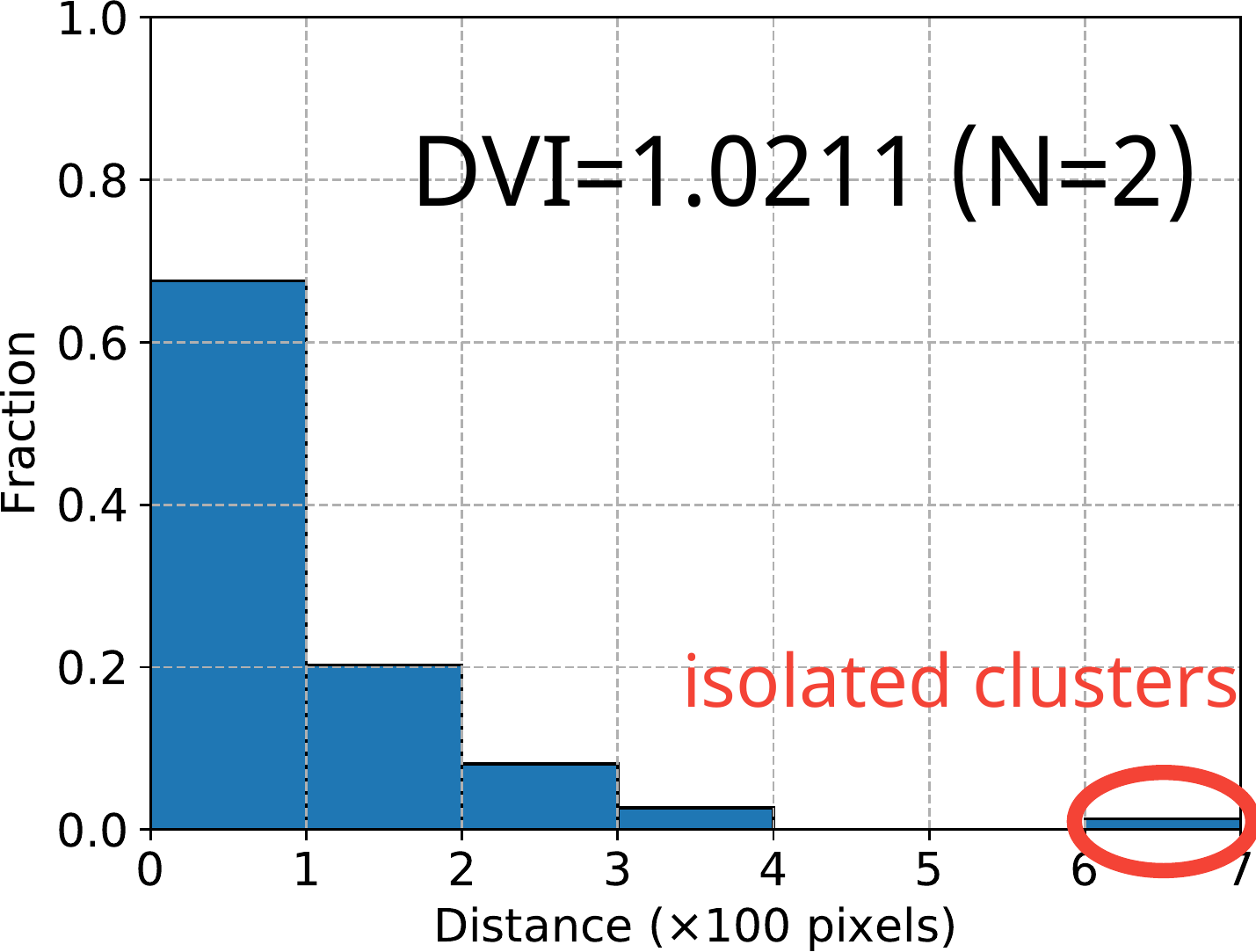} \\
		\hline
	\end{tabular}
	\label{table1}
	\vspace{-0.2in}
\end{table*}

\begin{itemize}
	\item {\em Large variation in object scale:} Due to the perspective and distance of the camera with respective to an object, two objects of the same physical size would appear as different size (or scale) in an image, the so-called 
	``scale variation.''  In a crowded scene, objects may have large scale variation.  
	Some works have studied this scale variation issue and achieved encouraging results using multi-branch architecture with different filter sizes~\cite{zhang2016single,sam2017switching}. However, they are often based on fixed kernel size, which is only sensitive within a few narrow bands
of receptive fields in the full spectrum, and hence cannot generally support diverse scenarios.
	
	\item {\em Isolated small clusters of objects:}  An image often has isolated object clusters, i.e.,
some targets are located far from the crowded mass.  These isolated clusters may be many but small in size, e.g., one or two objects in many scattered clusters.
	Convolutional operations with local receptive fields often cannot capture properly long-range contextual information, and hence have difficulty in modeling such isolated small clusters.  This leads to unsatisfactory results, especially for many such small isolated clusters.
\end{itemize}

Crowd scenes captured by drones or surveillance cameras
exhibit a wide range of scale variations and object clustering.  We
illustrate in Table~\ref{table1} three typical images with vehicles from the
VisDrone2019 dataset~\cite{DBLP:journals/corr/abs-1804-07437}.  We show the original images in the
first row; their corresponding density maps in the second; and the scale (object size)
distribution of the objects in pixels, in terms of the average of its length and
height, in the third row.
To understand the small clustering effect, we take the distances of the $N$ closest neighbors from an object and calculate their averages, for some small $N$ (say, 1-5).  A high average means that the object is in a small isolated cluster of size no more than $Nk$ from the mass, and vice versa.
In the fourth row of Table~\ref{table1}, we hence show the distribution of
the average distance of the nearest $N$ neighbors of an object, for $N =
2$.

We measure the degree of scale variation by the coefficient of variation (CV), defined as the ratio between the standard deviation and the mean of the object scale~\cite{abdi2010coefficient}.  
The larger the CV is, the more challenging the scale variation problem is.
Furthermore, we measure the level of small clustering by 
the Dunn Validity Index (DVI) using the distance distribution~\cite{maulik2002performance} after applying K-means algorithms~\cite{hartigan1979algorithm}
($K = 2$ in our case) on it.
The larger the DVI, the more small-size scattered clusters there are.

We show in Table~\ref{table1} the 
CV for scale variation and DVI for distance distribution for the three images.
The 
scale distribution of Image $A$ is wide, while its distance plot is rather continuous. Therefore, 
it is an image mainly characterized by scale variation without severe isolated clustering. This is validated with its relatively high CV on the scale (0.7592) and low DVI value (0.0205), and is clearly shown on its density map.
On the other hand,
the scale plot of Image $B$ shows a rather narrow distribution, while 
its distance plot contains some outliers (separate bars).
It is an image mainly with small isolated clusters, as validated with a low CV of scale (0.2054) and high DVI value (1.5871). Image $C$ is intermediate between the two images:
the size plot has a wide range and
the distance plot has outliers (separate bars).
It has
both scale variation and isolated clustering, with its CV of scale and DVI value between Images $A$ and $B$. This is clearly shown by its original image and density map.


To tackle diverse scale variation and capture long-range contextual information in isolated small clusters,
we propose \name{}, a 
{\bf s}cale-{\bf a}daptive long-range {\bf c}ontext-{\bf a}ware {\bf net}work 
for accurate crowd estimation and high-quality density map generation.
\name{} fully extracts the long-range contextual features in an image via a pyramid encoder by adaptively enlarging its receptive field. Using a novel scale-adaptive self-attention module with multi-branch architecture, it attains high scale sensitivity
and much better accuracy in 
detecting isolated small clusters (based on the extracted 
long-range contextual information).
Finally, it employs an efficient hierarchical fusion module to combine the multi-level scale and contextual features to generate a highly accurate density map. \name{} also uses group normalization to achieve higher optimality and better convergence by reducing small batch size influence.

We have conducted extensive experiments on
both drone-based and surveillance camera-based crowd datasets for unconstrained crowd counting. We use the images from the four datasets: VisDrone2019 People \& Vehicle, ShanghaiTech A \& ShanghaiTech B. As compared with the state-of-the-art approaches, \name{} achieves superior performance on all of the four challenging benchmarks.


The rest of the paper is organized as follows. In Section~\ref{related}, we review the related work. Then we present the details of \name{} in Section~\ref{scheme}, in terms of its pyramid contextual module, scale-adaptive self-attention module, hierarchical fusion, group normalization, and objective function. We discuss the experimental setting and illustrate the results in Section~\ref{results}, and conclude in Section~\ref{conclude}.

\section{Related Work}\label{related}

In this section, we discuss the related work of crowd counting methods in three main directions: traditional crowd counting algorithms (Section~\ref{tra}), deep learning-based approaches (Section~\ref{dl}), and crowd counting for drone-based scenes (Section~\ref{drone}).
\subsection{Traditional Approaches} \label{tra}
Early approaches for crowd counting are often based on detection models with hand-crafted features, i.e., they leverage pedestrian or body-part detectors to detect individual objects and count the number in the whole image~\cite{rabaud2006counting},~\cite{lin2010shape}. However, the performance of these detection-based methods degrades seriously in highly crowded scenes. Some researchers have attempted to use regression-based approaches with low-level features like HOG and SIFT to calculate the global number~\cite{chan2012counting}. Even though relying on low-level features, these approaches achieve better results for the global count estimation. To incorporate spatial information, researchers have proposed the density map regression-based approaches, that is, measuring the number of people per unit pixel of an area in a crowd scene.  As has been discussed in~\cite{lempitsky2010learning}, the work is the first one to provide a density map regression-based crowd counting approach with linear mapping algorithms. And then, another work improves it with random forest regression to learn non-linear mapping and achieves much better performance~\cite{pham2015count}.

\subsection{Deep Learning-based Approaches}   \label{dl}
Recently, researchers have adopted deep learning-based methods instead of relying on hand-crafted features to generate high-quality density maps and achieve accurate crowd counting~\cite{cao2018scale},~\cite{shen2018crowd}. These approaches can be applied to count different kinds of objects (i.e., vehicles and cells) instead of people~\cite{li2018csrnet},~\cite{he2019automatic}.

Researchers propose multi-column convolutional neural networks with different kernel sizes for each column to address the scale variation problem~\cite{zhang2016single}. Switching-CNN attaches a patch-based switching block to the multi-column structure, and better handles the particular range of scale for each column~\cite{sam2017switching}. HydraCNN utilizes a pyramid of image patches with multiple scales for crowd estimation~\cite{onoro2016towards}. However, the counting networks with inappropriate receptive field size will give an unbalanced focus to multi-scale targets. In addition, these methods only rely on static receptive field size, which cannot be extended to tackle wide-scale variation in a crowd scene. Besides, current approaches cannot fully leverage the long-range contextual information and have difficulty in modeling isolated small clusters.

\subsection{Drone-based Scenarios}   \label{drone}
Drone sensors and surveillance cameras both can capture crowd scenes, but crowd analysis for drone-based scenarios is more flexible for smart city applications~\cite{liu2019geometric}. However, previous works focus more on surveillance camera-based crowd counting. To the best of our knowledge, drone-based crowd counting has not yet been fully explored, and it also lacks publicly large-scale diversified datasets. Besides, compared with images taken by surveillance cameras, the isolated small clusters problem is more severe for drone-based crowd images~\cite{qiu2019local}. In this paper, we modified the VisDrone2019 challenging dataset~\cite{zhuvisdrone2018} into two large-scale diversified drone-based crowd counting datasets, and these newly split datasets can promote the field. Besides, we tackle the two main challenging scale variation and the isolated small clusters problem for unconstrained crowd counting within an end-to-end framework.

\begin{figure*}[t]
	\centering
	\vspace{-0.15in}
	\includegraphics[width=0.92\textwidth, height=0.48\textwidth]{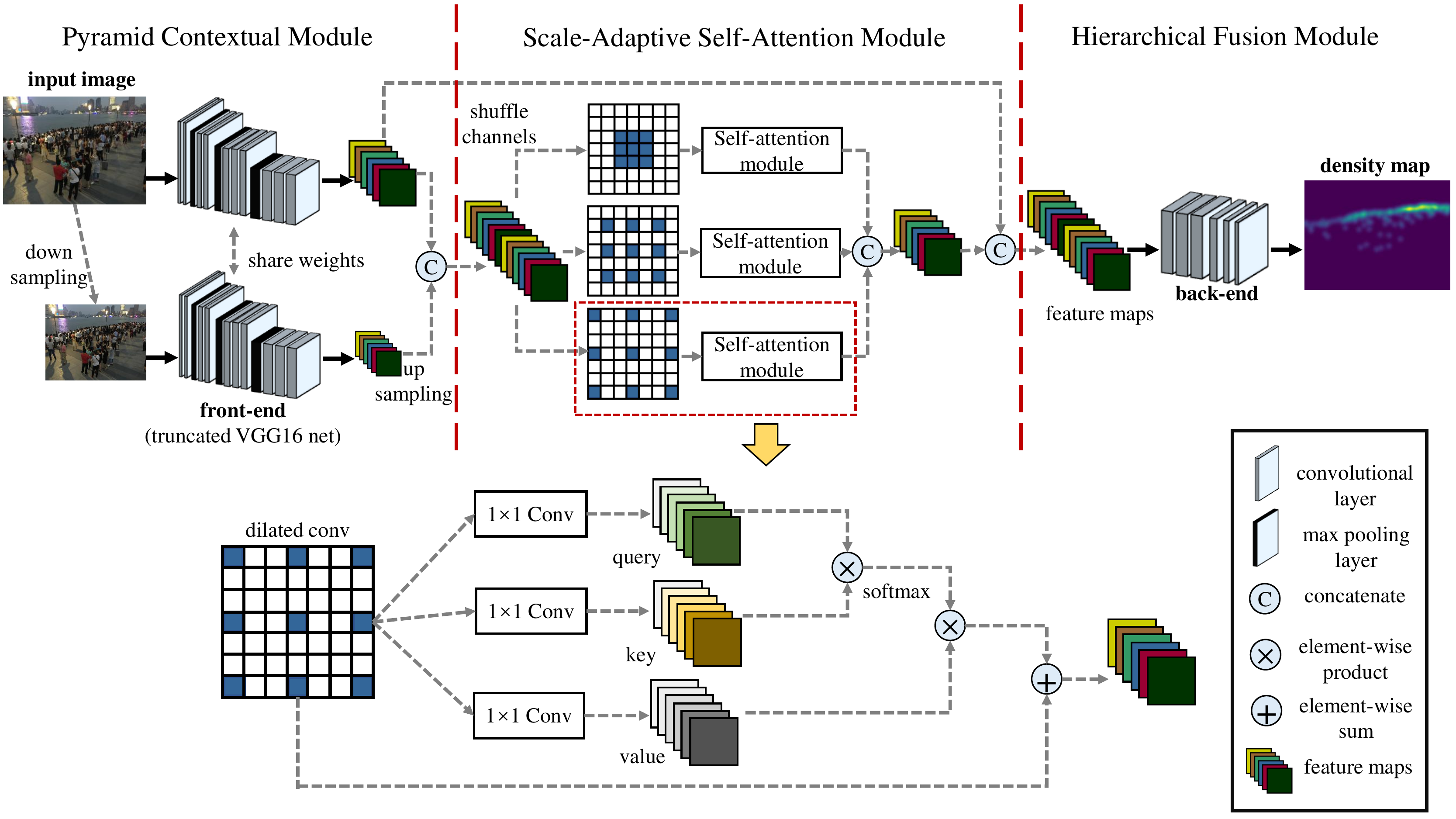}
	\caption{The architecture of \name{} for crowd counting. It contains three main components: the pyramid contextual module, the scale-adaptive self-attention module, and the hierarchical fusion module. 
        }
	\label{image1}
	\vspace{-0.20in}
\end{figure*}

\section{SACANet for Crowds with Scale Variation and Isolated Clusters} \label{scheme}

In this section, we present the details of \name{}, a novel {\bf s}cale-{\bf a}daptive long-range {\bf c}ontext-{\bf a}ware {\bf net}work  for crowd counting.
\name{} consists of three components: the pyramid contextual module (Section~\ref{pcm}), the scale-adaptive self-attention module (Section~\ref{sasam}), and the hierarchical fusion module (Section~\ref{hfm}). 
Besides, we describe how to utilize group normalization to facilitate convergence and achieve better optimality in Section~\ref{gn}. The details of the objective function is discussed in Section~\ref{loss}. 


In Figure~\ref{image1}, we show the architecture of \name{}. The upper part presents the three components of \name{}, and we describe the scale-adaptive self-attention module in more details in the bottom part (under the yellow arrow).   


\subsection{Pyramid Contextual Module} \label{pcm}

The pyramid contextual module aims to extract long-range contextual information and enlarge the receptive field. For fair comparison with previous works~\cite{li2018csrnet}~\cite{chen2019scale}~\cite{liu2019point}, our approach incorporates truncated VGG-16~\cite{simonyan2014very} with excellent transferability as the backbone. We extract the first ten layers of VGG-16 with only three pooling layers to balance the large valid receptive field and density map resolution. Besides, adopting multi-branch only for the higher layers can benefit memory efficiency during training.

The pyramid contextual module is composed of a feature extractor and contextual fusion. We downsample the input image to $1/4$ of the original resolution and then feed these two kinds of inputs to the VGG-16 front-end with shared weights. We shuffle the channels of the generated feature maps before feeding them into the second scale-adaptive self-attention module. This component can enlarge the receptive field and capture long-range contextual information. The encoder part is also designed to facilitate the downstream scale-adaptive self-attention processing. 

\subsection{Scale-Adaptive Self-Attention Module}\label{sasam}

The scale-adaptive self-attention module is to accommodate wide scales and detect isolated clusters via the long-range contextual information. This part consists of three branches with the same filter size ($3\times3$), but different dilated ratios~\cite{yu2015multi} ($1\times1$, $2\times2$, $3\times3$). We also add a separate convolution layer with a filter size $1\times1$ at the beginning of each branch in order to reduce the numbers of channels to $1/4$ of its input. This can help to reduce memory requirements without sacrificing performance~\cite{szegedy2017inception}. Besides, the scale-adaptive self-attention module is one of the key elements of each branch after the dilated convolutional layer.

The simplest way to use multi-branch features is to concatenate them~\cite{zhang2016single}.
However, the features with different receptive fields would be quite large and may contain redundant information. Inspired by~\cite{vaswani2017attention}, we utilize a scale-adaptive self-attention module to capture long-range dependencies, which computes a weighted sum of values and assigns weights to measure the importance of this branch. Crowd analysis suffers from scale variations. A single multi-branch structure only possesses the same weight for each branch, and inappropriate filter size has a bad effect on the estimation. Our scale-adaptive self-attention module is able to decide for itself which to focus.

The scale-adaptive self-attention module first transfers input $x$ to query $Q_x$, key $K_x$ and value $V_x$:
\vspace{-0.1in}
\begin{equation}
Q_x = f(x), K_x = g(x), V_x = h(x).
\vspace{-0.1in}
\end{equation}

The output weighted density map $Y$ is computed by two kinds of matrix multiplications:
\vspace{-0.1in}
\begin{equation}
Y = \textup{softmax} \left (Q_xK_x^{T} \right )V_x.
\vspace{-0.1in}
\end{equation}

Therefore, our scale-adaptive self-attention module can automatically choose the most suitable branches and enlarge the receptive field with limited extra parameters.

\subsection{Hierarchical Fusion Module}              \label{hfm}

The hierarchical fusion module integrates multi-level self-attention features to achieve accurate crowd counting estimation. It can take advantage of autofocusing branches with different receptive fields and generate high-quality density maps. We need long-range contextual information from the deeper layers with a large receptive field and semantic information. At the same time, we also require short-range contextual information for each unit. Therefore, our method can accurately model multi-level contextual information and recognize isolated small clusters.

Inspired by ResNet~\cite{he2016deep} for recognition and RefineNet~\cite{lin2017refinenet} for semantic segmentation, we gradually refine all the details of the generated density maps from the previous layers. Besides, there is a fully convolutional network backbone in the end to recover the spatial information. Finally, an output convolutional layer is used to predict the density map value for each pixel. Note that ReLU operations are added after each convolutional layer~\cite{glorot2011deep}, and the entire network can be efficiently trained end-to-end.

\subsection{Group Normalization for Better Convergence}   \label{gn}

When directly training the entire network, we find that the model cannot converge well due to a gradient vanishing problem. We tried batch normalization~\cite{ioffe2015batch} but the result is not ideal because the error increases when the batch size becomes smaller. Our task cannot use a large batch size due to large image resolution and limited computation memory. Inspired by~\cite{wu2018group}, we utilize group normalization instead of batch normalization in our approach for better convergence. 

GN separates the channels into different groups and calculates the mean $\mu$ and standard deviation $\sigma$ within each group. This has no relation to batch size and enables our network to converge. The formulation for the mean and standard deviation are as follows:
%
$\mu_i=\frac{1}{m}\sum_{k\in S_i}x_k, \; \sigma_i=\sqrt{\frac{1}{m}\sum{k\in S_i}(x_k -\mu_i)^2+\epsilon}$,
%
where $\epsilon$ is a small constant, and m is the size of the set. $S_i$ is the set of pixels where the mean and standard deviation are computed. For group normalization, set $S_i$ can be defined as, $S_i={k|k_{N}=i_{N}, \; \lfloor{\frac{k_C}{C/G}}\rfloor=\lfloor{\frac{i_C}{C/G}}\rfloor}$,
%
%
where G is the number of groups, which is a hyper-parameter we need to decide. $C/G$ is the number of channels for each group, and $\lfloor \cdot \rfloor$ is the floor operation. In our experiment, we set the channels per group at 16 if the total number of channels is larger than 16, or we let G be the same as the number of channels.

\subsection{Objective Function}     \label{loss}

Most of the recent works use Euclidean loss to optimize their models for crowd counting~\cite{zhang2016single}, we also use it to optimize the aforementioned network. The Euclidean loss is a pixel-wise estimation error, which is defined as: 
\vspace{-0.12in}
\begin{equation}
L_{E}=\frac{1}{N} \left ||F(X;\alpha)-Y \right ||_{2}^{2},
\vspace{-0.12in}
\end{equation}
where $\alpha$ indicates the model parameters, N means the number of pixels, X denotes the input image and Y is its ground truth and $F(X;\alpha)$ is the generated density map. We can predict the crowd counting result by summarizing over the estimated crowd density map.



\section{Experiments and Illustrative Results}\label{results}

In this section, we describe the evaluation metrics and comparison schemes in Section~\ref{metrics}. The description of the four datasets and ground truth generation method is presented in Section~\ref{datasets}. Section~\ref{train} shows our training details. Qualitative and quantitative analysis of both people and vehicle datasets are detailed in Section~\ref{visdrone}. Besides, we conduct ablation study in Section~\ref{as} and compare SACANet on unconstrained scenarios in Section~\ref{us}. 


\subsection{Evaluation Metrics and Comparison Schemes } \label{metrics}

We use the coefficient of variation (CV) to measure the degree scale variation~\cite{abdi2010coefficient}. For the level of isolated small clusters, we leverage the Dunn Validity Index (DVI) to calculate the distance distribution after applying K-means algorithms on it~\cite{maulik2002performance}. The two evaluation metrics are defined as follows: $\mathrm{CV} = {\sigma}/{\mu}$,
where $\sigma$ is the standard deviation, and $\mu$ is the mean,
\vspace{-0.1in}
\begin{equation}
\mathrm{DVI} = \frac{\min \limits_{0<m \neq n<K} \{ \min \limits_{\forall x_i \in \Omega_m,\forall x_j \in \Omega_n} \{ ||x_i - x_j|| \}\} }{\max \limits_{0<m \neq n \le K}\max \limits_{\forall x_i,x_j \in \Omega_m}\{ || x_i - x_j || \}}.
\vspace{-0.1in}
\end{equation}

DVI calculates the shortest inter-cluster distance divided by the maximum inner-cluster distance. In our experiments, we apply the $K$-means algorithm~\cite{hartigan1979algorithm} ($K=2$ in our experiment) to preprocess the average $N=2$ nearest distance data, and then calculate the DVI value. The larger the DVI value, the larger the inter-cluster distance and the shorter the inner-cluster distance, and the more isolated small-size clusters there are.

For overall crowd counting results, two metrics are used for evaluation~\cite{willmott2005advantages}, Mean Absolute Error (MAE) and Mean Squared Error (MSE), which are defined as follows:
\vspace{-0.1in}
%
\begin{equation}
\mathrm{MAE} = \frac{1}{N}\sum_{i=1}^{N}|C_{i}-\hat{C}_{i}|, 
\mathrm{MSE} = \sqrt{\frac{1}{N}\sum_{i=1}^{N}|C_{i}-\hat{C}_{i}|^{2}},
\vspace{-0.12in}
\end{equation}
%
where $N$ is the total number of test images, $C_{i}$ means the ground truth count of the $i$-th image , and $\hat{C}_{i}$ represents the estimated count.

\begin{table*}
	\centering
	\caption{Statistics of different datasets in our experiment.}
	\begin{tabular}{|c|c|c|c|c|c|}
		\hline
		Dataset       &Average Resolution &Images &Max &Min &Total \\
		\hline\hline
		VisDrone2019 People~\cite{zhuvisdrone2018} & 969$\times$1482 &3347 &289 &10  &108,464 \\
		VisDrone2019 Vehicle~\cite{zhuvisdrone2018} & 991$\times$1511 &5303&349 &10  &198,984\\
		ShanghaiTech A~\cite{zhang2016single} & 589$\times$868   &482    &313   &33  &241,677\\
		ShanghaiTech B~\cite{zhang2016single} & 768$\times$1024   &716   &578   &9   &88,488\\
		\hline
	\end{tabular}	
	\label{table2}
	\vspace{-0.22in}
\end{table*}


We compare our approach with three schemes on the VisDrone2019 datasets: VGG-16 ~\cite{simonyan2014very}, MCNN~\cite{zhang2016single} and CSRNet~\cite{li2018csrnet}. VGG-16 is a strong backbone, and we directly modify it into a crowd counting network. Multi-Column Convolutional Neural Network (MCNN) is a well-known crowd counting approach with multi-branch architecture. We implement it in our framework with three branches, which is the same as ours for a fair comparison.
CSRNet is one of the state-of-the-art methods for congested scenes understanding, which leverages dilated convolution to enlarge the receptive field. We implement it with the same experiment settings as ours.



\begin{figure*}[t]
	\centering
	\includegraphics[width=1\textwidth, height=0.54\textwidth]{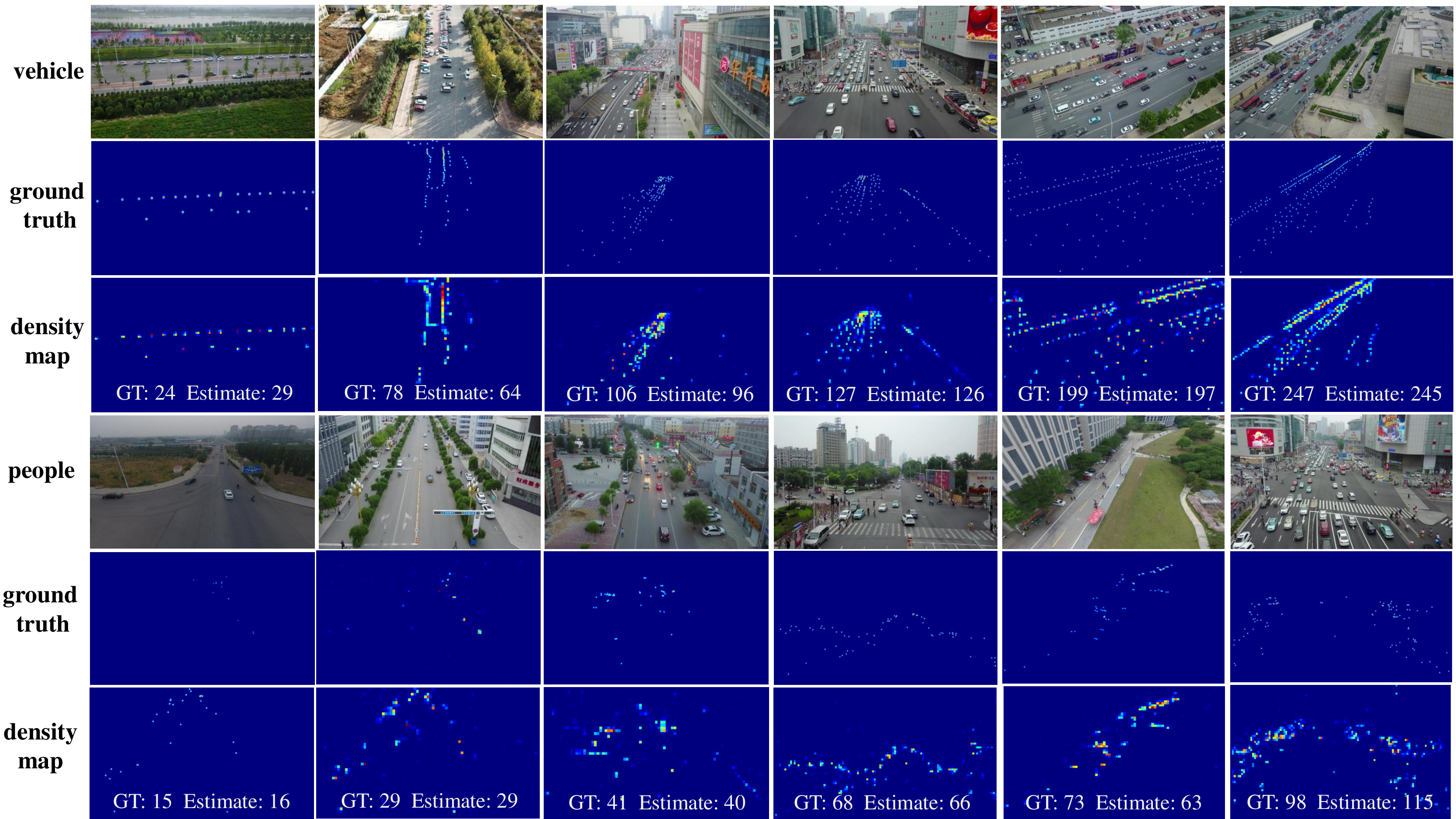}
	\caption{Qualitative results of SACANet on the VisDrone2019 Vehicle \& People dataset.}
	\label{image2}
	\vspace{-0.05in}
\end{figure*}

\begin{table*}
	\begin{minipage}{\linewidth}
		\begin{minipage}{0.46\linewidth}
			\centering
			\caption{Quantitative results on the VisDrone2019 Datasets.}\label{table3}
			\begin{tabular}{|c|c|c|c|c|}
				\hline
				Method &\multicolumn{2}{|c|}{Dataset People} &\multicolumn{2}{|c|}{Dataset Vehicle}\\
				\cline{2-5}
				~ &          MAE &MSE &       MAE &MSE \\
				\hline\hline
				VGG-16 ~\cite{simonyan2014very}            &22.0 &44.8 &21.4 &29.3 \\
				MCNN ~\cite{zhang2016single}            &16.4 &39.1 &14.9 &21.6 \\
				CSRNet  ~\cite{li2018csrnet}          &12.1 &36.7 &10.9 &16.6 \\
				\hline
				SACANet(ours)&$\textbf{10.5}$&$\textbf{35.1}$&$\textbf{8.6}$&$\textbf{12.9}$\\
				\hline
			\end{tabular}
		\end{minipage}
		\begin{minipage}{0.56\linewidth}
			\centering
			\caption{Ablation study on the VisDrone2019 dataset.}\label{table4}
			\begin{tabular}{|c|c|c|c|c|}
				\hline
				Method &\multicolumn{2}{|c|}{Dataset People} &\multicolumn{2}{|c|}{Dataset Vehicle}\\
				\cline{2-5}
				~ &          MAE &MSE &       MAE &MSE \\
				\hline\hline
				baseline             &14.5 &40.8 &12.5 &19.1 \\
				baseline+context      &13.8 &39.2 &11.0 &16.4\\
				baseline+context+SASA &11.7 &36.1 &9.1 &13.6\\
				\hline
				SACANet(ours)&$\textbf{10.5}$&$\textbf{35.1}$&$\textbf{8.6}$&$\textbf{12.9}$ \\
				\hline
			\end{tabular}
		\end{minipage}
	\end{minipage}
	\vspace{-0.2in}
\end{table*}

\begin{figure*}[t]
	\centering
	\includegraphics[width=0.495\textwidth, height=0.265\textwidth]{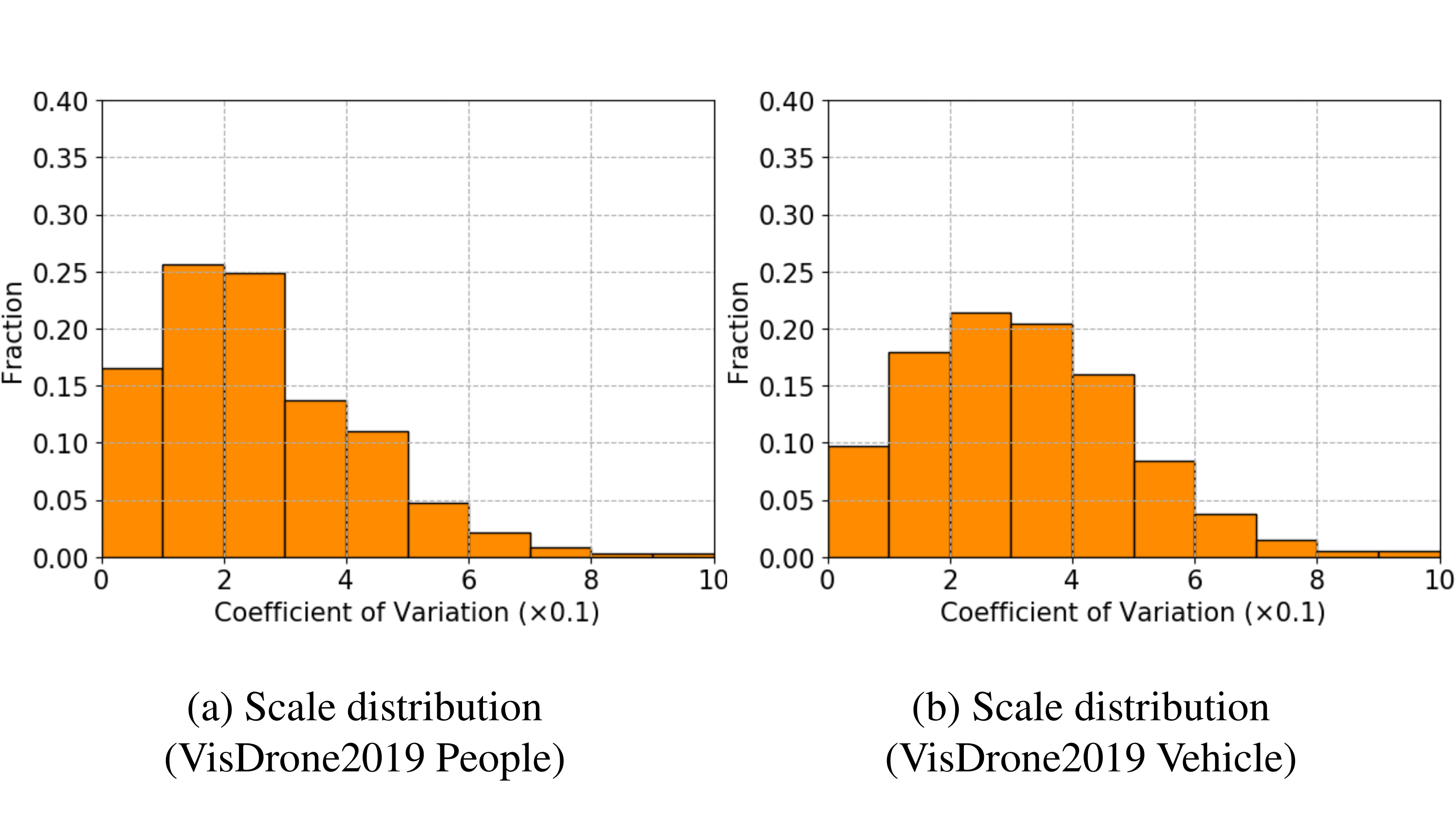}
	\includegraphics[width=0.495\textwidth, height=0.265\textwidth]{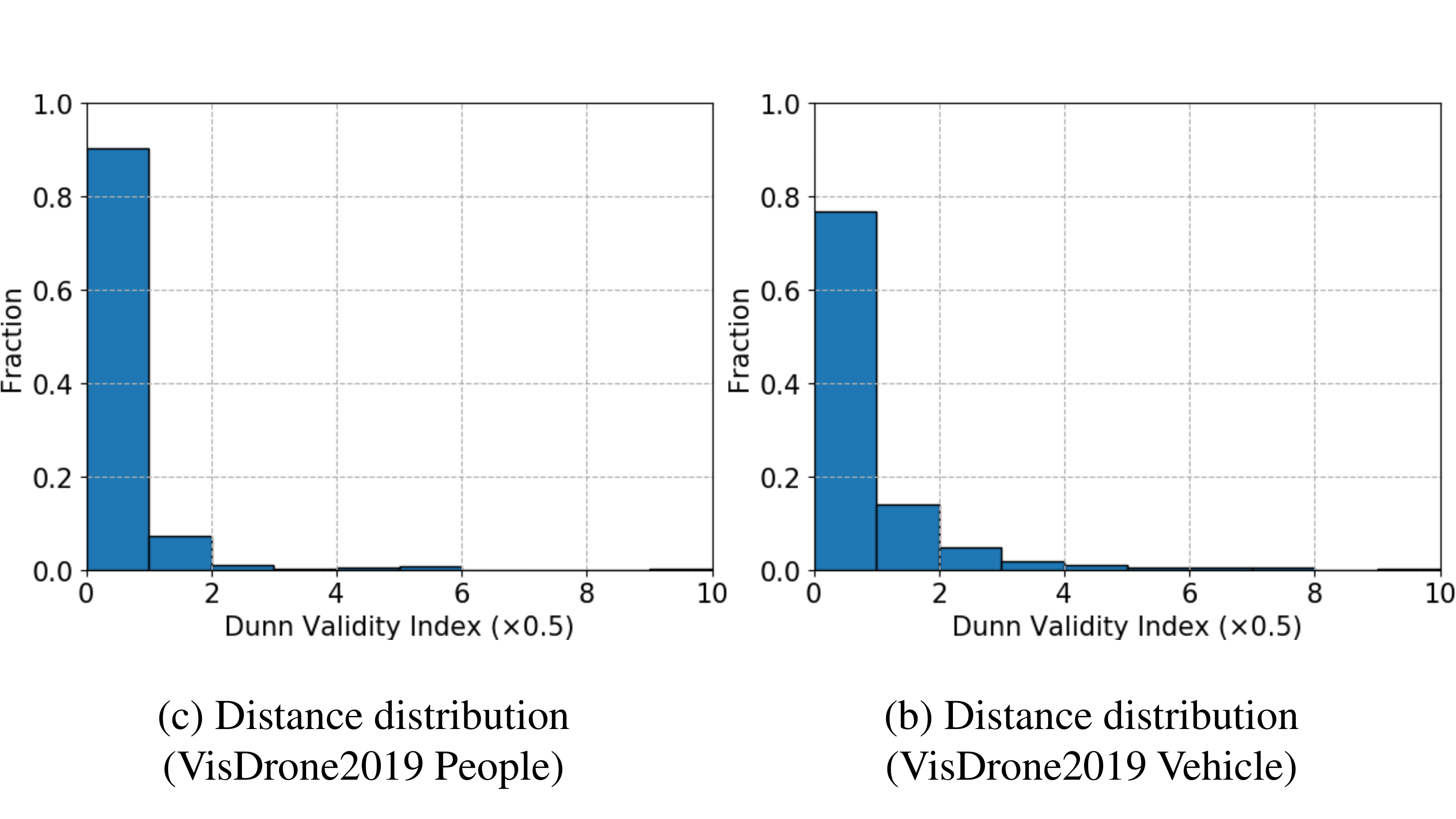}
	\caption{Distribution of scale variation and object separation problem on the VisDrone2019 Vehicle \& People datasets.}
	\label{image3}
	\vspace{-0.15in}
\end{figure*}

\begin{figure*}[t]
	\centering
	\includegraphics[width=0.495\textwidth, height=0.265\textwidth]{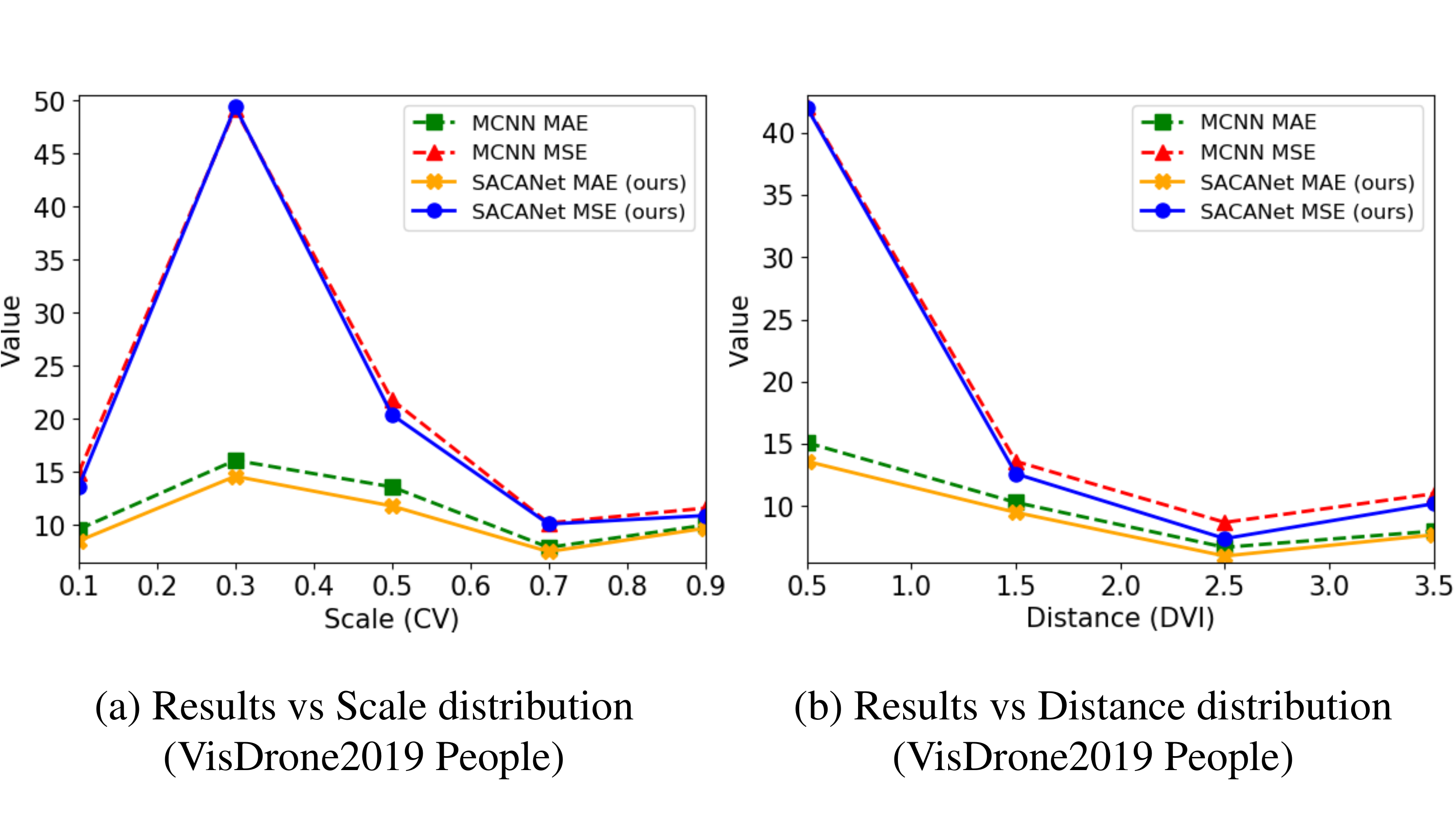}
	\includegraphics[width=0.495\textwidth, height=0.265\textwidth]{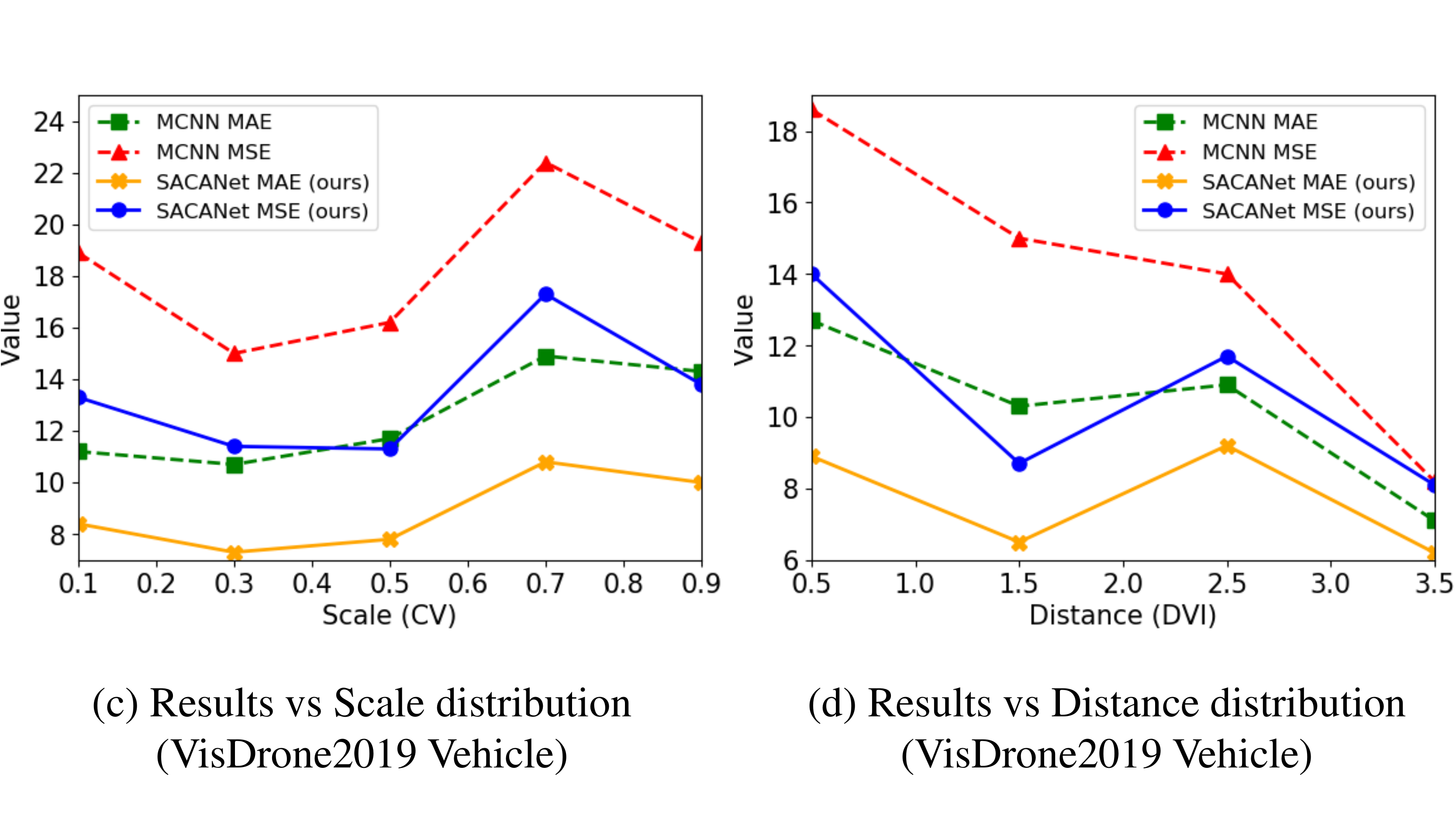}
	\caption{Comparison via different distributions on the VisDrone2019 Vehicle \& People datasets. }
	\label{image4}
	\vspace{-0.15in}
\end{figure*}

\subsection{Datasets and Ground Truth Generation}  \label{datasets}
We evaluate our method on four challenging crowd counting datasets: VisDrone2019 People \& Vehicle, and ShanghaiTech A \& B. The statistics of the four datasets are presented in Table~\ref{table2}.

VisDrone2019 People~\cite{DBLP:journals/corr/abs-1804-07437}. We modify the original VisDrone2019 object detection dataset~\cite{zhuvisdrone2018} with bounding boxes of targets to crowd counting annotations. The original VisDrone2019 dataset contains 11 categories. Category 0 (pedestrian) and category 1 (people) are combined into one dataset for people crowd counting, and the new annotation location is the head point of the original bounding box. We rearrange the data split and filter out the cases whose number of annotated target objects is less than 10. Finally, this dataset consists of 2392 training samples, 329 validation samples, and 626 test samples.

VisDrone2019 Vehicle~\cite{DBLP:journals/corr/abs-1804-07437}. Following the similar modification and rearranging step for the VisDrone2019 People dataset, we combine category 4 (car), category 5 (van), category 6 (truck) and category 9 (bus) into one dataset for vehicle crowd counting. We get 3953 training samples, 364 validation samples, and 986 test samples. The new vehicle annotation location is the center point of the original bounding box. These two kinds of new annotation operations are defined as follows:
\vspace{-0.1in}
\begin{equation}
\begin{aligned}
\mathrm{People[cols, rows]}&=\mathrm{[bb_{left}+\frac{bb_{width}}{2}, bb_{top}]}, \\
\mathrm{Vehicle[cols, rows]}&=\mathrm{[bb_{left}+\frac{bb_{width}}{2}, bb_{top}+\frac{bb_{height}}{2}]}.
\end{aligned}
\vspace{-0.05in}
\end{equation}

ShanghaiTech A~\cite{zhang2016single}. The ShanghaiTech A dataset includes 482 crowd images with a total number of 241, 677 persons. The counts for each image vary from 33 to 3139. This dataset is randomly crawled from the Internet and has unfixed resolutions.

ShanghaiTech B~\cite{zhang2016single}. The ShanghaiTech B dataset has 716 images with fixed resolutions, which were taken from busy shopping streets by fixed cameras. The counts for each image vary from 12 to 578.

We generate the ground truth by way of blurring each head or center of vehicle annotation with a Gaussian kernel (which is normalized to 1) and taking the spatial distribution of all images into consideration from each dataset. For the sparse crowd dataset VisDrone2019 People, VisDrone2019 Vehicle, and ShanghaiTech B, we use fixed kernel $\sigma$ = 15 as the generating method. For ShanghaiTech A crowded datasets, we adopt the geometry-adaptive kernels method to generate the ground truth~\cite{zhang2016single}.

\subsection{Training Details} \label{train}

In our experiment, we use the VGG-16 backbone with pretrained weights from the ImageNet classification challenge dataset~\cite{krizhevsky2012imagenet}. For the other layers, we initialize with random weights from Gaussian distributions with zero mean and a standard deviation of 1. We use the Adam optimizer~\cite{kingma2014adam} with an initial learning rate of 1e-4. For the ShanghaiTech B dataset, we randomly flip some of the samples to augment the original training data. For datasets with much higher resolutions, we resize the input image to no more than $768\times1024$ but maintain the same aspect ratio.

In the test step, we directly test the full images for the ShanghaiTech B dataset with fixed resolutions. For the other datasets with unfixed resolutions, we resize the images before feeding them into our network. The estimated total count for each image is given by summing the whole image. We implement SACANet based on PyTorch. 



\subsection{Experiments on the VisDrone2019 Datasets} \label{visdrone}

We modify the original VisDrone2019 challenge dataset into two crowd counting datasets (VisDrone2019 Vehicle \& People) and conduct extensive experiments on the two new datasets. Details are presented in Table~\ref{table2}. We implement a strong VGG-16 network and the state-of-the-art approaches MCNN~\cite{zhang2016single}, and CSRNet~\cite{li2018csrnet} and compare with SACANet on the same dataset. SACANet achieves much better performance on both the VisDrone2019 People and the VisDrone2019 Vehicle datasets, and the quantitative results are presented in Table~\ref{table3}. 

Figure~\ref{image2} shows the qualitative results on both the VisDrone2019 Vehicle and the VisDrone2019 People datasets. The first three rows are the results for vehicle crowd counting, and the last three rows are results for people crowd counting. For each part, the first row shows the original image of the VisDrone2019 dataset, the second row presents the ground truth, and the third row is our generated density map. Besides, the total number of ground truth and our estimated counting results are shown below each part.

Furthermore, we split the VisDrone2019 Vehicle \& People datasets based on the different levels of scale variation and object separation problem distribution. In Figure~\ref{image3}, we plot perception vs. the evaluation metrics (i.e., Coefficient of Variation for object size distribution and Dunn Validity Index for distance distribution). Figure~\ref{image3} (a) and (b) presents the scale distribution measured by CV. (c) and (d) shows the distance distribution measured by DVI.

For the scale variation problem, we divided the test dataset into five parts based on the value of CV. Thus, we get a five-level scale variation problem setting with different CV ranges: [0.0,$\ $0.2), [0.2,$\ $0.4), [0.4,$\ $0.6), [0.6,$\ $0.8), [0.8,$\ \infty$). The higher the scale split set level, the more difficult the scale variation problem. Besides, for the isolated small clusters problem, the test dataset is decomposed into four subsets based on the DVI value, which indicates four levels of the isolated clusters problem: [0.0,$\ $1.0), [1.0,$\ $2.0), [2.0,$\ $3.0), [3.0,$\ \infty$). The higher the distance split sets the level, the more small-size isolated clusters there are.

We compared SACANet with MCNN~\cite{zhang2016single} on the split subsets. In Figure~\ref{image4} (a) and (c), we plot the value of MAE and MSE vs. scale (CV) ranges. For (b) and (d), we plot the value MAE and MSE vs. distance (DVI) ranges. The results of SACANet are always better than MCNN on most of the subsets. Our method achieves much lower MAE on the VisDrone2019 Vehicle than on the people dataset as the vehicle dataset is relatively more crowded, which indicates that SACANet is more effective for extremely crowded scenes.

\subsection{Ablation Study} \label{as}

In this section, we perform ablation studies on the VisDrone2019 People \& Vehicle datasets and analyze the results, which shows the effectiveness of our approach.

Effectiveness of the baseline: Our baseline network consists of three parts: a truncated VGG-16, multi-branch subnet, and the backend. For a given input image, we feed it to our baseline network and get its generated density map. And then, we sum all the pixel values to get the total count.
	
Effectiveness of the pyramid context module: After the operations mentioned above, we use our context-aware front-end instead of one column VGG-16 backbone to train our model, and we find that the error decreased.
	
Effectiveness of the Scale-Adaptive Self-Attention (SASA): We enrich the baseline with our novel scale-adaptive attention scheme to each branch. The results show a significant improvement in terms of MAE and MAE.
		
Effectiveness of the hierarchical fusion: We add the hierarchical fusion module and retrain the model. This module also brings improvement to the performance, and we get our final results in Table~\ref{table4}.

\begin{table}
	\centering
	\caption{Results of our approaches on two challenging people counting datasets. ShanghaiTech A \& B.}
	\begin{tabular}{|c|r|r|r|r|}
		\hline			
		Method &\multicolumn{2}{|c|}{Dataset A} &\multicolumn{2}{|c|}{Dataset B}     \\
		\cline{2-5}
		~ &          MAE &MSE &       MAE &MSE \\
		\hline\hline
		MCNN ~\cite{zhang2016single}    &110.2 &173.2 &26.4  &41.3  \\
		Switching-CNN ~\cite{sam2017switching}              &90.4 &135.0 &21.6   &33.4  \\
		CP-CNN ~\cite{sindagi2017cnn}                     &73.6  &106.4  &20.1   &30.1  \\
		ACSCP ~\cite{shen2018crowd}                      &75.7  &102.7 &17.2   &27.4  \\		
		IG-CNN ~\cite{babu2018divide}                     &72.5  &118.2 &13.6   &21.1  \\		
		CSRNet ~\cite{li2018csrnet}                     &68.2  &115.0  &10.6   &16.0  \\
		DRSAN ~\cite{liu2018crowd}                      &69.3   &96.4  &11.1   &18.2   \\
		SANet ~\cite{cao2018scale}                      &67.0  &104.5  &8.4 & 13.6 \\
		\hline
		Baseline              &68.3 &107.5 &11.9   &18.7    \\
		SACANet(ours)               &$\textbf{64.4}$ &$\textbf{95.9}$  &$\textbf{7.8}$   &$\textbf{13.5}$    \\
		\hline
	\end{tabular}
	\label{table5}
	\vspace{-0.2in}
\end{table}

\subsection{Evaluation on Unconstrained Scenarios}  \label{us}

We compare SACANet with other approaches in the literature on two challenging people crowd counting datasets with unconstrained scenarios. The results are proposed in Table~\ref{table5}. Our methods always show a better performance than the baseline, which demonstrates the importance of the pyramid contextual module and scale-adaptive self-attention mechanism. Besides, we see that our approach surpasses the state-of-the-art methods CSRNet~\cite{li2018csrnet} and SANet~\cite{cao2018scale} for both the ShanghaiTech A and the ShanghaiTech B datasets. Compared with SANet~\cite{cao2018scale}, our SACANet reduces the MAE by over 7\% on the ShanghaiTech A dataset and reduces the MAE by about 4\% on the ShanghaiTech B dataset. These results further demonstrate the effectiveness of our \name{}.

\section{Conclusion}\label{conclude}


In this work, we tackle two main challenges in crowd counting: large scale variation and isolated small clusters. We propose \name{}, a novel Scale-Adaptive long-range Context-Aware network for accurate crowd counting in unconstrained crowded scenes. A pyramid contextual module can fully encode the contextual information. We present a scale-adaptive self-attention scheme to automatically choose the most appropriate branches and naturally enlarge the receptive field. By utilizing the hierarchical fusion module, our method can fuse multi-level contextual information in crowded scenes.

Extensive experiments show that our approach achieves compelling results on the VisDrone2019 Vehicle \& People datasets and two other challenging people crowd counting benchmarks. As compared with prior arts, \name{} achieves much better performance in terms of MAE and MSE, especially when the image exhibits large variation in object scales and many isolated small clusters.


{\small
\bibliographystyle{ieee_fullname}
\bibliography{egbib}
}

\end{document}